\documentclass[10pt,twocolumn,letterpaper]{article}
\usepackage[utf8]{inputenc}
\usepackage{graphicx} 
\usepackage{bm}
\usepackage{amsmath}
\usepackage{amssymb}
\usepackage{listings}
\usepackage{booktabs}
\usepackage{url}

\begin{document}
\title{ \LARGE \bf Face Recognition in the age of CLIP \&  Billion image datasets}
\author{Aaditya Bhat\\
Independent Scholar\\
{\tt\small aadityaubhat@gmail.com}
\and
Shrey Jain\\
Department of Electrical \& Computer Engineering, New York University\\
{\tt\small sj3396@nyu.edu}
}

\date{}
\maketitle

\begin{abstract}

CLIP (Contrastive Language-Image Pre-training) models developed by OpenAI have achieved outstanding results on various image recognition and retrieval tasks, displaying strong zero shot performance. This means that they are able to perform effectively on tasks for which they have not been explicitly trained. Inspired by the success of OpenAI CLIP, a new publicly available dataset called LAION-5B was collected which resulted in the development of open ViT-H/14, ViT-G/14 models that  outperform the OpenAI L/14 model. The LAION-5B dataset also released an approximate  nearest neighbor index, with a web interface for search and subset creation.

In this paper, we evaluate the performance of various CLIP models as zero shot face recognizers. Our findings show that CLIP models perform well on face recognition tasks, but increasing the size of the CLIP model does not necessarily lead to improved accuracy. Additionally, we investigate the robustness of CLIP models against data poisoning attacks by testing their performance on poisoned data. Through this analysis, we aim to understand the potential consequences and misuse of search engines built using CLIP models, which could potentially function as unintentional face recognition engines. 

\end{abstract}

\section{Introduction}
CLIP (Contrastive Language-Image Pre-training) is a type of deep learning model developed by OpenAI that has achieved impressive results on a variety of image recognition and retrieval tasks. The CLIP model was introduced in “Learning Transferable Visual Models From Natural Language Supervision“ by Radford et. al. (https://arxiv.org/abs/2103.00020), which describes the training of the model on a proprietary dataset of 400 million images collected from the web. The models were publicly released in 2021 and can be found on the OpenAI GitHub page.

CLIP is trained by combining a large amount of text and image data and using an objective function that encourages the model to map images and text to a shared latent space. This enables CLIP to perform well on tasks such as image classification and retrieval, even when it has not been explicitly trained on those tasks.

One of the key characteristics of CLIP is its ability to perform zero shot learning, meaning it can recognize and classify objects or concepts that it has not seen before. This is achieved through the use of a shared latent space for text and images, which allows the model to generalize from the text data to the image data. For example, if CLIP has been trained on a large dataset of images and their associated textual descriptions, it can use its understanding of the words in the description to make educated guesses about the contents of a new image.Overall, the development of CLIP has significantly advanced the field of image recognition and retrieval, and it has the potential to be applied to a wide range of tasks and industries.

The CLIP model was trained on a dataset of 400 million (image, text) pairs that were collected from various publicly available sources on the internet. To ensure a diverse range of visual concepts was included in the dataset, the creators of CLIP searched for (image, text) pairs using a set of 500,000 queries and included up to 20,000 (image, text) pairs per query. The resulting dataset was called WIT for WebImageText. This dataset is proprietary and wasn’t released with the models.

LAION-5B dataset was created to address the lack of publicly available datasets with billions of image-text pairs, which are necessary for training powerful multimodal learning models like CLIP, DALL-E, etc. The LAION-5B dataset provides researchers with a publicly available resource for training and testing these types of models. LAION-5B is a large dataset of image-text pairs for use in language-vision research. The dataset consists of 5.85 billion image-text pairs, with 2.32 billion containing English language. LAION-5B can be used to replicate and fine-tune foundational models and to perform additional experiments.

In addition to the image-text pairs, the LAION-5B dataset also includes CLIP ViT-L/14 embeddings, kNN-indices, tools for NSFW and watermark detection, and a web interface for exploration and subset creation. LAION-5B was collected by parsing files in the Common Crawl dataset to find image tags with alt-text values. The corresponding images were downloaded and filtered using CLIP to keep only those images whose content resembled their alt-text description. The LAION-5B dataset offers a valuable resource for research on multi-modal language-vision models and is available to the broader research community.

CLIP models have demonstrated strong transfer learning and out-of-distribution generalization capabilities, making them potential candidates for use as facial recognizers. Even though CLIP was not specifically trained for facial recognition, it has been shown to be able to extract rich facial features (Goh et al., 2021) and is highly resistant to image perturbations (Radford et al., 2021). Fine-tuned CLIP models have also been found to be robust against existing data poisoning attacks when used as facial recognizers (Radiya-Dixit et al., 2022). A. Mart´ı and V. Rodriguez-Fernandez  used the  CLIP model  to construct a natural language-based version of the game "Guess who?" in which players engage with the game using language prompts and CLIP determines whether or not an image meets the prompt. The performance of this technique is evaluated using various question prompts, and the limitations of its zero-shot capabilities are demonstrated.

In this paper, we evaluate the performance of various CLIP models as zero shot face recognizers, examining how well they are able to accurately identify and classify different faces. We also investigate the robustness of CLIP models against data poisoning attacks, testing their performance on poisoned data to understand the potential consequences and misuse of these models as face recognizers. Through this analysis, we aim to understand the potential of CLIP models as tools for face recognition and the potential consequences of using these models in real-world applications.

\begin{figure*}[t]
\includegraphics[width=\textwidth, height=8cm]{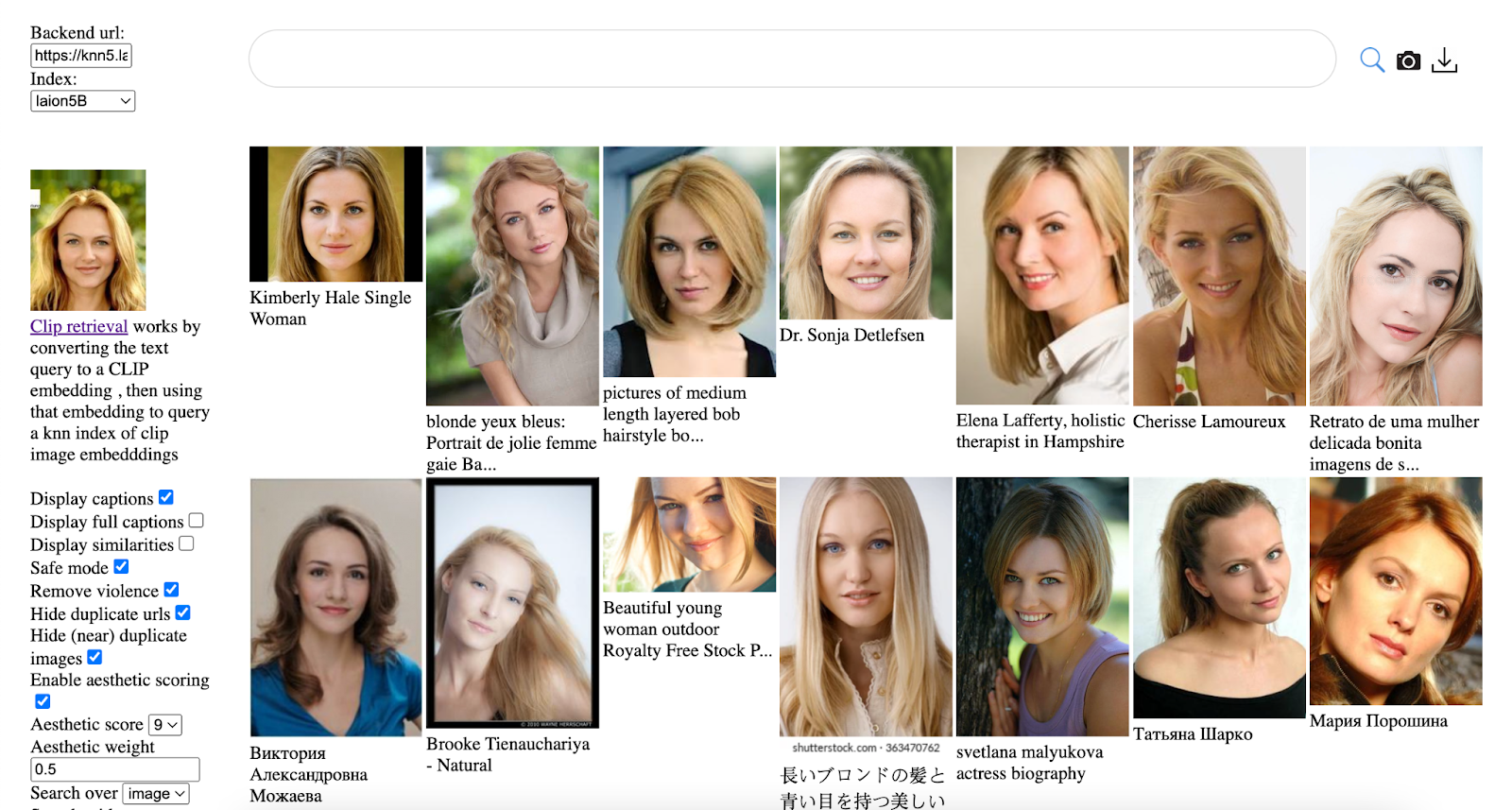} 
\caption{CLIP-retrieval web interface}
\label{fig:figure1}
\end{figure*}

\section{Experiments}
In this experiment, we evaluate the performance of various CLIP models as zero-shot face recognizers using the CelebFaces Attributes (CelebA) Dataset as our primary dataset. This dataset contains 202,599 face images of various celebrities, with 10,177 unique identities, and is diverse in facial features, expressions, and demographics. We calculate top-1 and top-5 face recognition accuracies of ViT-B/32, ViT-L/14, ViT-H/14, ViT-G/14 models on this dataset and compare them to the top-1 and top-5 face recognition accuracy of the face-recognition python library. Further, we conduct data poisoning attacks on a subset of 1000 images from the CelebA dataset using Fawkes and LowKey tools, and use the unperturbed, Fawkes cloaked, and LowKey attacked versions of the images as query images to search the LAION-5B KNN index. We analyze the images retrieved for the original, Fawkes, and LowKey versions to investigate the robustness of the CLIP models against data poisoning.

\subsection{CLIP Accuracy Experiment}
We compute the ViT-B/32, ViT-L/14, ViT-H/14, ViT-G/14, and face-recognition embeddings for 202k images in the dataset. We build a flat L2 index over the embeddings for each model. For each image, we measure the L2 distance between the query image embeddings and all the embeddings loaded into the index. We retrieve the top 6 images with the minimum L2 distance from the index. As the query image will always be present in the index, the image with second minimum distance will be the top-1 image, and [2,6] images from top 6 will be the top-5 images.  Our evaluation metrics are top-1 accuracy and top-5 accuracy.

\begin{figure*}[t]
\includegraphics[width=\textwidth, height=9cm]{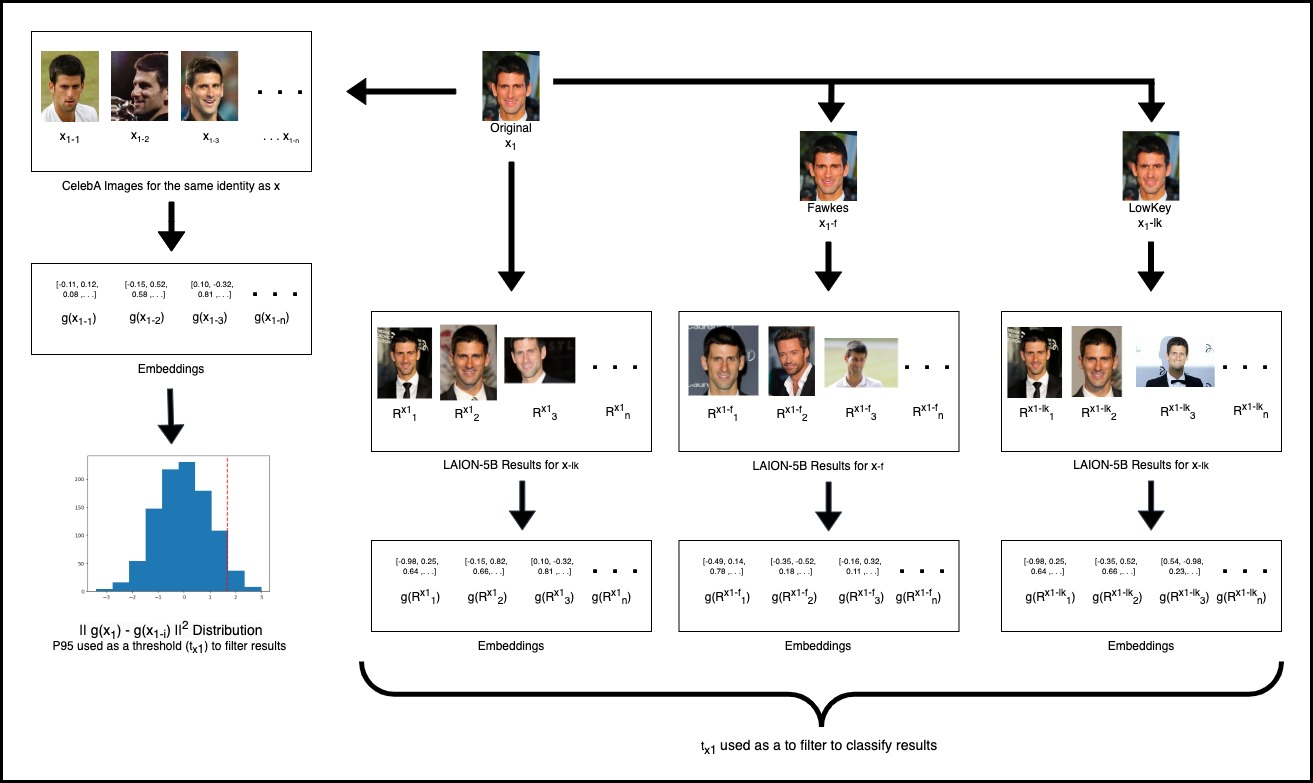} 
\caption{Face Recognition Capabilities of Image Retrieval Engines Experiment Setup }
\label{fig:figure2}
\end{figure*}

\subsection{Face Recognition Capabilities of Image Retrieval Engines }

We take a subset of 1000 images from the CelebA dataset. For each image $(x_n)$ in the subset, we compute the distribution of L2 distance between the query image $(x_n)$ and all other images of the same identity $(x_{n-i})$ from the CelebA dataset. We take the 95th percentile of this distribution as the max distance threshold $(t_{xn})$.

Further, for each image$(x_n)$ in the subset, we create the Fawkes perturbed version($x_n$-f) and LowKey perturbed version($x_n$-lk) of it. The Fawkes version of images are created using CLI provided at (https://github.com/Shawn-Shan/fawkes) with protection mode set to mid. The LowKey version of images are created using Python code provided with the paper.

For each version, original, Fawkes and LowKey, we search the LAION-5B KNN index to retrieve up to 50 images ($R^{xn}_1$ , $R^{xn}_2$ , $R^{xn}_3$ , … $R^{xn-f}_n$ ; $R^{xn-f}_1$ , $R^{xn-f}_2$ , $R^{xn-f}_3$ , … $R^{xn-f}_n$ ; $R^{xn-lk}_1$ , $R^{xn-lk}_2$ , $R^{xn-lk}_3$ , … $R^{x-lk}_n$ ;) The various search parameters for are set as follows:
\begin{lstlisting}
    url ="https://knn5.laion.ai//knn-service",
    indice_name = "laion5B",
    use_mclip = False,
    aesthetic_score = 9,
    aesthetic_weight = 0.5,
    modality = Modality.IMAGE,
    num_images = 50,
    deduplicate = True,
    use_safety_model = False,
    use_violence_detector = False,
\end{lstlisting}

We filter out the results $R^{xn}_1$ , $R^{xn}_2$ , $R^{xn}_3$ , … $R^{xn-f}_n$ ; $R^{xn-f}_1$ , $R^{xn-f}_2$ , $R^{xn-f}_3$ , … $R^{xn-f}_n$ ; $R^{xn-lk}_1$ , $R^{xn-lk}_2$ , $R^{xn-lk}_3$ , … $R^{x-lk}_n$ ;) where the $|| g(x_n) - g(result) ||^2  > t_{xn}$ . This limits the results to valid results, i.e.  result images with the same identity as the query image. We further limit the subset images to images with 3 or more valid results for the original version of the image. Among these images, we look at how many valid results did Fawkes and LowKey versions returned.

\section{Results}

From the CLIP accuracy experiment, we see that CLIP models perform with varying degrees of success at face recognition. For top-1 prediction, ViT-L/14 model has the highest accuracy among CLIP models of 80.95\%, which is still significantly worse than the face-recognition python package, which has 87.61\% accuracy. For top-5 prediction, ViT-H/14 model has the highest accuracy among  CLIP models of 89.88\%, which is comparable to 92.27\% accuracy of face-recognition python package. Face recognition accuracy of CLIP models does not always improve with the bigger models. For both, top-1 and top-5 predictions, we see that the accuracy initially improves and then deteriorates as models get bigger.

\begin{figure}[ht]
\includegraphics[width=0.5\textwidth]{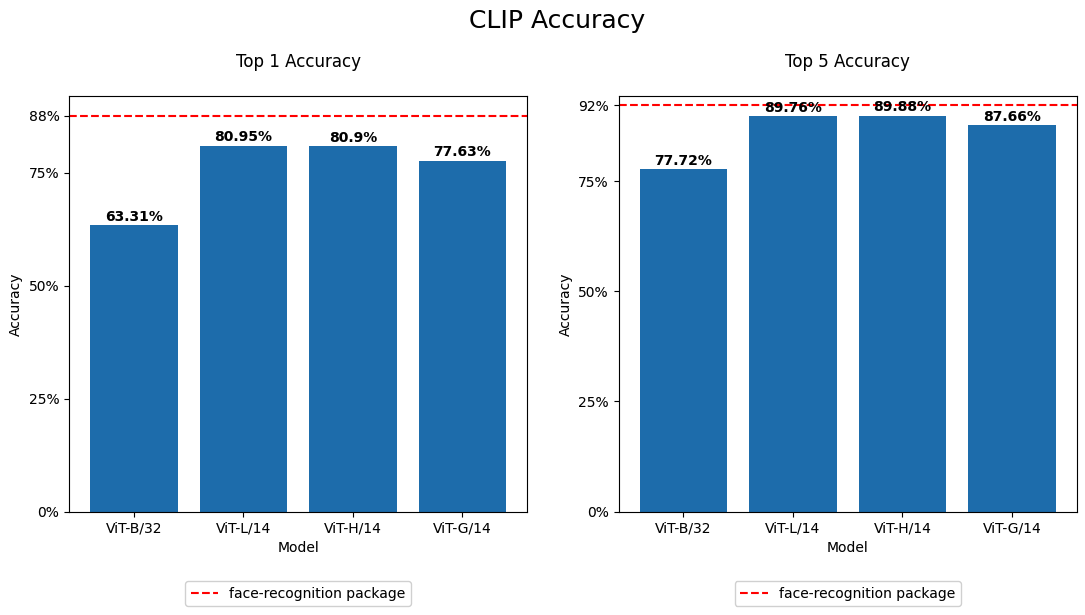} 
\caption{CLIP Face Recognition Accuracy}
\label{fig:figure3}
\end{figure}

For the Face Recognition Capabilities of Image Retrieval Engines experiment, we see that out of 1000 images from the subset, 728 images return 3 or more valid results when queried with the original version of the image on the LAION-5B KNN index. Among these 728 images, 612 (84.07\%) images return one or more valid results when queried with the Fawkes perturbed version of the image, and 566 (77.75\%) images return one or more valid results when queried with the LowKey perturbed version of the image. Among the 728 images, 543 (74.59\%) images return one or more valid results for both Fawkes and LowKey perturbed versions.

\section{Discussion}
Biometric Information Privacy Act (BIPA) passed by Illinois, and similar laws passed by Texas and Washington, regulate the collection, use, and handling of biometric identifiers and information by private entities. BIPA defines biometric identifiers as a retina or iris scan, fingerprint, voiceprint, or scan of hand or face geometry. This definition is broad, and one can argue that CLIP vectors can constitute as biometric identifiers, as CLIP vectors can be used for facial recognition. If CLIP vectors are considered as biometric identifiers, using them for facial recognition would require compliance with BIPA and similar laws passed by other states.

Additionally, the potential for misuse of CLIP search engines as inadvertent face recognition engines is a serious concern. Even if the intended use of the search engine is not for facial recognition, it could still be used for that purpose. This raises important questions about the responsibility of the creators of these models to ensure that their use does not violate privacy and civil rights.

Given the potential for misuse, it would be valuable to explore the possibility of training new CLIP models to intentionally have worse performance on face recognition tasks. However, this approach has its own limitations and challenges such as it will not solve the problem of unintended use or abuse of the technology, it might also be difficult to ensure that the models are not used for nefarious purposes despite the intended use.

\section{Conclusion}
CLIP models show good zero-shot face recognition capabilities and are robust against data poisoning attacks. Increasing the size of the CLIP model does not necessarily lead to improved accuracy in face recognition tasks. CLIP models have demonstrated impressive performance on image recognition and retrieval tasks, it is essential to consider the potential consequences and misuse of search engines built using these models, which could inadvertently function as face recognition engines.The use of CLIP models for face recognition raises a number of important legal and ethical questions that need to be addressed.

\end{document}